\documentclass[10pt, conference, compsocconf]{IEEEtran}
\usepackage{amssymb}
\usepackage{graphicx}
\usepackage{float} 
\usepackage{subcaption}
\usepackage{tabularx}
\usepackage{xstring}
\usepackage{multirow}
\usepackage{amsmath}

\ifCLASSINFOpdf
\else
\fi
\begin{document}
%
\title{Efficient Unsupervised Video Object Segmentation Network Based on Motion Guidance}


\author{\IEEEauthorblockN{1st Chao Hu\textsuperscript{*}}
\IEEEauthorblockA{AI Lab\\
Unicom (Shanghai) Industry Internet Co., Ltd.\\
Shanghai, China\\
huchao.000@gmail.com}
\and
\IEEEauthorblockN{2nd Liqiang Zhu}
\IEEEauthorblockA{AI Lab\\
Unicom (Shanghai) Industry Internet Co., Ltd.\\
Shanghai, China\\
zhulq0@gmail.com}
}


%


\maketitle

\begin{abstract}
Due to the problem of performance constraints of unsupervised video object detection, its large-scale application is limited. In response to this pain point, we propose another excellent method to solve this problematic point. By incorporating motion characterization in unsupervised video object detection, detection accuracy is improved while reducing the computational amount of the network. The whole network structure consists of dual-stream network, motion guidance module, and multi-scale progressive fusion module. The appearance and motion representations of the detection target are obtained through a dual-stream network. Then, the semantic features of the motion representation are obtained through the local attention mechanism in the motion guidance module to obtain the high-level semantic features of the appearance representation. The multi-scale progressive fusion module then fuses the features of different deep semantic features in the dual-stream network further to improve the detection effect of the overall network. We have conducted numerous experiments on the three datasets of DAVIS 16, FBMS, and ViSal. The verification results show that the proposed method achieves superior accuracy and performance and proves the superiority and robustness of the algorithm.

\end{abstract}

\begin{IEEEkeywords}
unsupervised video object segmentation; dual-stream network; motion representation; appearance representation; fusion module

\end{IEEEkeywords}

%
\IEEEpeerreviewmaketitle

\section{Introduction}
Unsupervised video object segmentation (UVOS) purposes to automatically gain salient objects from videos without human intervention. This task of automatically segmenting primary objects has recently received extensive attention and has significantly impacted all aspects of computer vision, involving safety monitoring, industrial manufacturing, autonomous driving, robotics.

Traditional methods usually use handcrafted features to address this problem, such as motion boundaries \cite{ref1}, sparse representations \cite{ref3}, salient \cite{ref2,ref5}, and point trajectories \cite{ref3,ref4,ref7}. Although the above algorithms have achieved inevitable success, they are not ideal for accurately discovering the objects which are most salient from the entire video sequence. With the rise of deep learning, several recent studies have attempted to model this problem as the zero-target frame problem \cite{ref8,ref9}. These studies basically learn object features from a large number of model training data, and these data are not manually labeled. Although the above methods have made breakthroughs, there are still problems. With the heavyweight networks, better feature representations can be extracted, such as the DeepLabv3 network \cite{ref26} based on the ResNet101 network. Capturing salient objects with a complex mechanism will lead to enormous model parameters, a higher model calculation amount, and a slower model training and inference speed, limiting the algorithm's application in practical situations.

How to efficiently capture significant objects is the key to network lightweighting. Recent studies have used the co attention mechanism to capture similar objects \cite{ref10} between different video frames, which has achieved good results but cannot distinguish objects similar to significant targets in the background, and the calculation is extensive. The motion information can capture salient objects based on human sensitivity to moving targets. At the same time, due to the prior information on the slow-moving objects in the video, the motion information extraction method based on local matching is more efficient.  So this paper uses the optical flow estimation network to extract motion information.

\begin{figure*}[t!]
	\centering
    \includegraphics[width=0.9\textwidth]{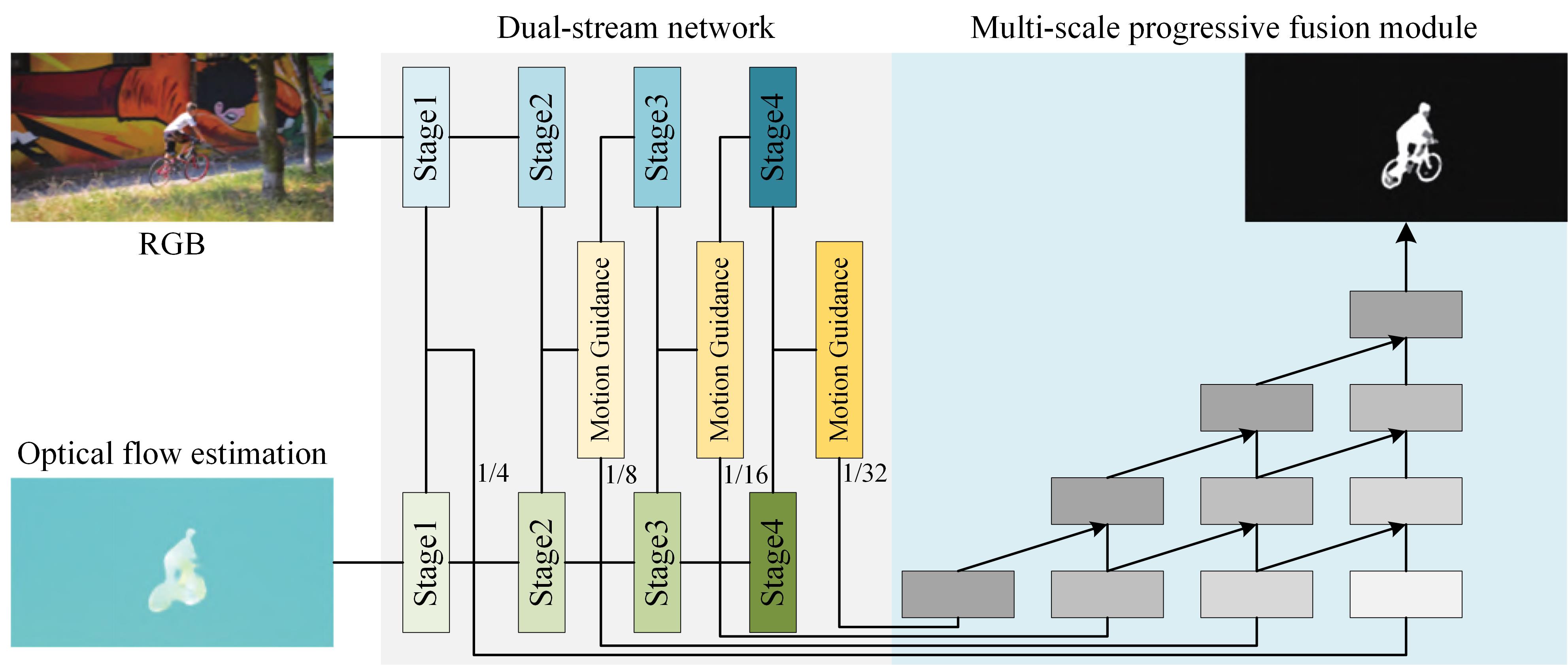}
	\caption{\centering{Network framework diagram}}
	\label{fig:1}
\end{figure*}

 Simultaneously, the appearance features of the objects in the RGB image are extracted to supplement the specific details lacking in the motion information and improve the final segmentation effect. Due to RGB images have a pixel correspondence with optical flow estimation, the motion information in optical flow estimation also includes the approximate position and contour information of salient objects. As a consequence, local attention mechanism can be utilized in motion information to obtain the convolution weights to guide Appearance features, learn semantics and reduce the difficulty of feature extraction in the RGB image branch. This method of motion information-guided appearance information learning enables the algorithm in this paper to obtain good feature extraction quality while using a lightweight feature extractor, and the amount of model parameters and model calculation are effectively reduced. Finally, multi-stage features are extracted to fed into the multi-scale progressive fusion module. The semantic information of high-resolution features is continuously enhanced through convolution and upsampling, and more accurate segmentation results are obtained. This article mainly has three contributions, which can be summarized as follows.
 \begin{itemize}
\item A lightweight method about unsupervised video object segmentation is proposed, which significantly reduces the model parameters and computational cost, and significantly improves the speed of the unsupervised video target segmentation algorithm.
\item A motion guidance module based on local attention force is proposed with prior motion information. Semantic information is extracted from motion information by local attention, and appearance features are guided to learn semantics information in convolutional weights, ultimately improving the segmentation performance.
\item Compared with the current SOTA methods, the results of our method is competitively on multiple standard datasets, proving the proposed algorithm's effectiveness and achieving a balance between speed and accuracy.
\end{itemize}

\section{Related work}
\subsection{Unsupervised Video Object Segmentation}
Early UVOS methods usually analyze point trajectories \cite{ref3,ref4,ref7} and use object suggestions \cite{ref11}, motion boundaries \cite{ref1}, or salient information \cite{ref2,ref5} to infer targets. However, the effect is not ideal due to the limitations of data sets and computing power. Thanks to establishing large datasets \cite{ref12,ref13} and developing fully convolutional segmentation networks, there have been various models are developed utilizing zero-target frame solutions to solve the problem.

\begin{figure*}[t!]
    \centering
    \begin{subfigure}[t]{0.4\textwidth}
           \centering
           \includegraphics[width=\textwidth]{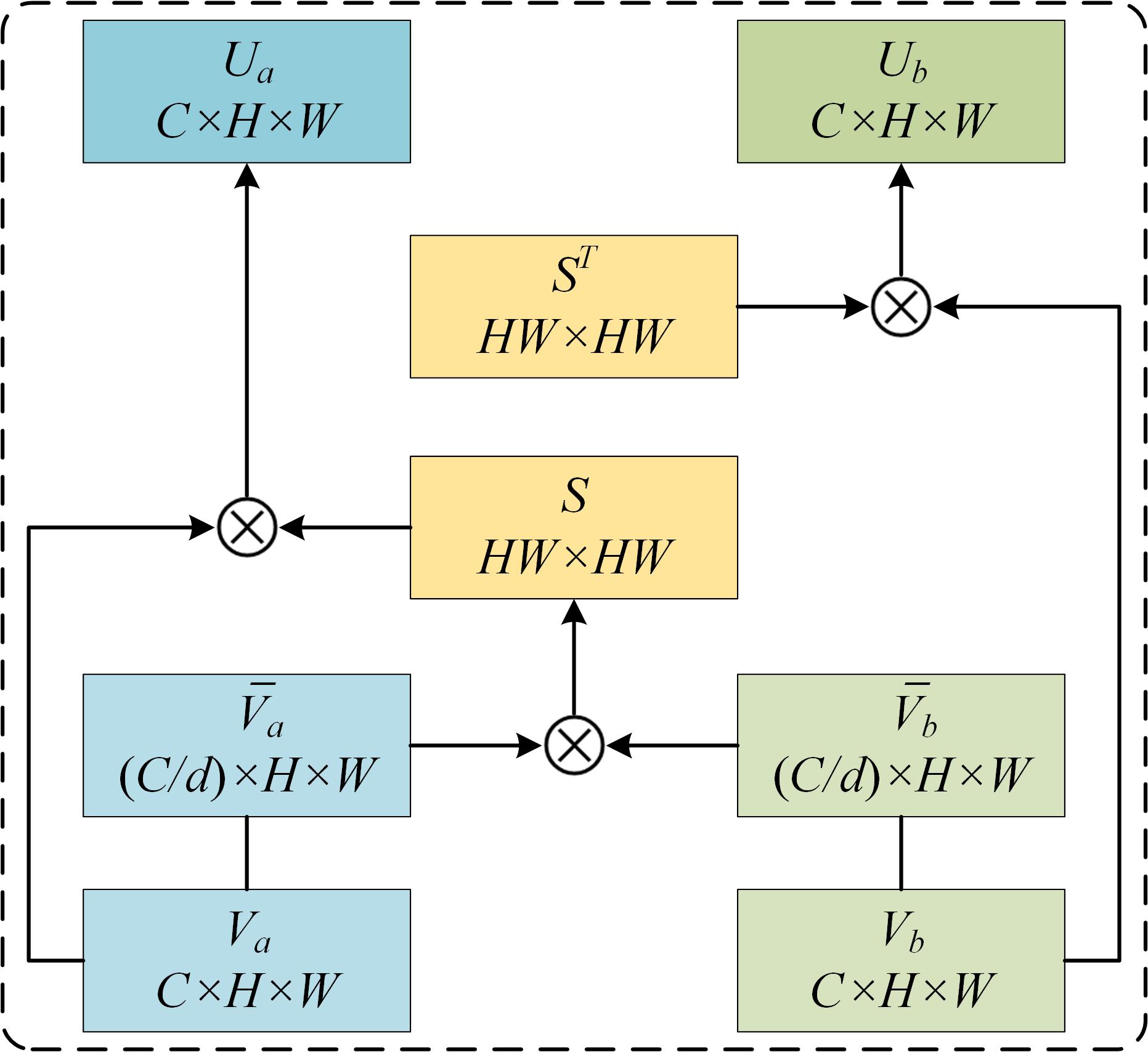}
            \caption{Co attention module}
            \label{fig:2_1}
    \end{subfigure}
    \begin{subfigure}[t]{0.5\textwidth}
            \centering
            \includegraphics[width=\textwidth]{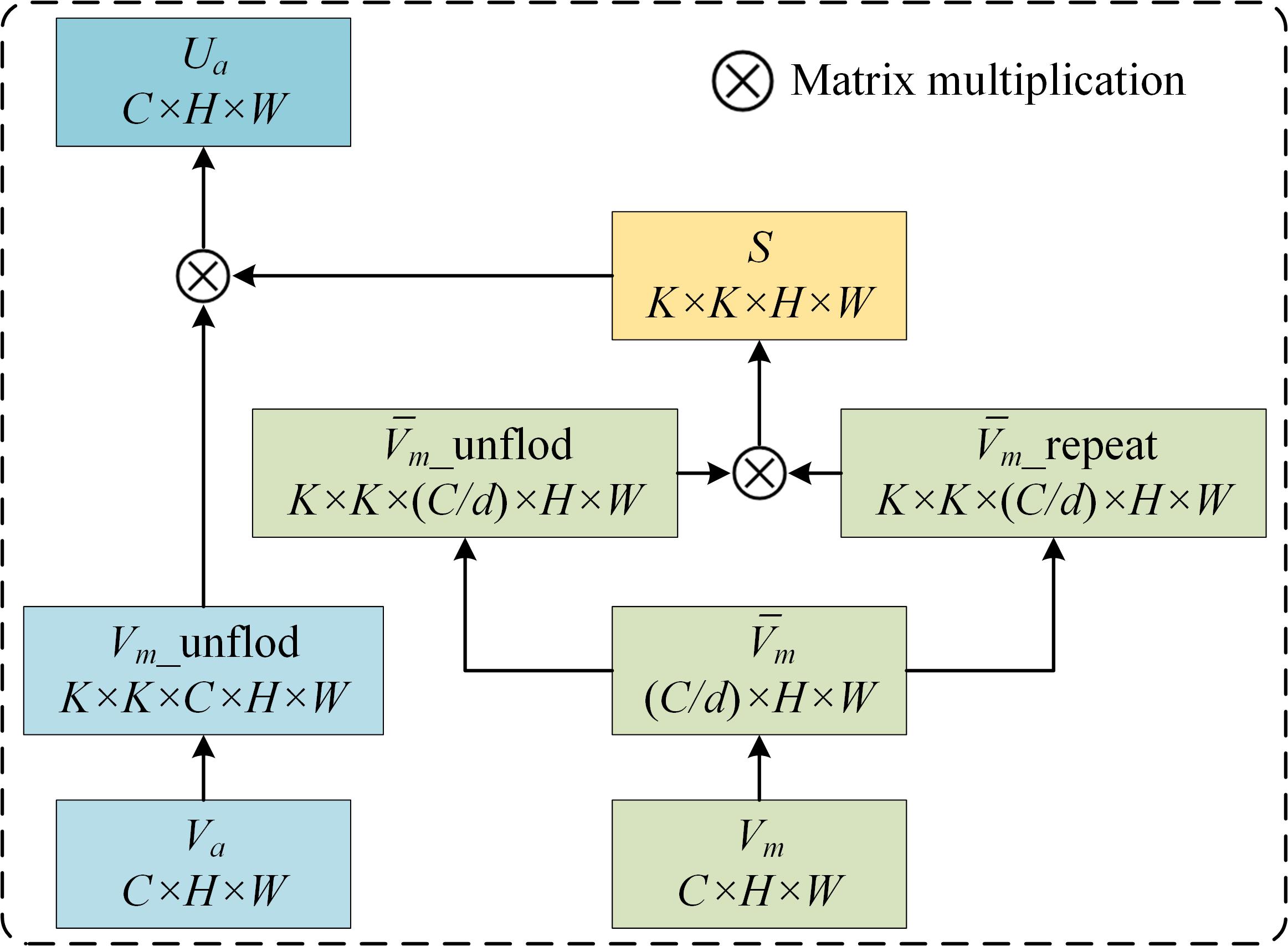}
            \caption{Motion guidance module}
            \label{fig:2_2}
    \end{subfigure}
    \caption{Illustration of attention structure}
\end{figure*}

One way to salient segment objects are to detect salient objects by video \cite{ref14}. In this way, the pre-trained semantic segmentation network will be fine-tuned, extracts salient features spatially, and then trains a convolutional long short-term memory (Conv-LSTM) to capture temporal information. With the emergence of the attention mechanism, for the sake of obtaining spatial and temporal information between different video frames for reasoning, the related methods use twin networks with co attention mechanism \cite{ref10}. However, it cannot sufficiently distinguish objects similar to salient objects in the background. Dual-stream networks \cite{ref16,ref17} are also popular, fusing motion with appearance information for objective reasoning. For example, MATNet \cite{ref15} uses the co attention mechanism to fuse motion and appearance features between various stages of the dual-stream network to obtain salient features, which has achieved relatively good results. However, the co attention mechanism in these studies brings massive computational costs, limiting the application in practical scenarios.

\subsection{Co Attention Mechanism}
Inspired by human perception, attention mechanism have been extensively researched and developed. Through end-to-end training, the model has ability to selectively focus on the subset of inputs by attention mechanism. For example, multi-context information is used to estimate human posture \cite{ref29}, and the spatial and channel attention is utilized to dynamically select a patch of  an image as the image description \cite{ref30}. Lately, co attention mechanism in linguistic and visual fields have been studied, such as visual question answering \cite{ref31} and visual conversations \cite{ref32}. In these works, the co attention mechanism is applied to explore the potential relevance among different patterns. For instance, a model was created in a previous visual question-answering study \cite{ref31}, jointly performing problem-guided visual and image-guided problem attention.  So the trained model can selectively pay attention to image patches and document fragments. Inspired by these works, the attention model in this paper is used to mine information among features with prior information, with a more elegant network architecture to capture motion information and guide appearance information to learn salient features.

\section{PROPOSED METHOD}
The overall network framework diagram of this article proposed is shown in Fig. \ref{fig:1}, which consisting mainly of three parts: dual-stream network, motion guidance, and multi-scale progressive fusion.

\subsection{Dual-stream Network}
In this article, the dual-stream network is constructed to  to extract motion and appearance features, which have proven its superiority in the video object detection task. Different from previous studies, this paper uses lightweight networks to replace DeepLabV3 networks \cite{ref26} based on ResNet101 and inserts motion guidance modules at different stages of lightweight networks to enhance the semantics of appearance features. This paper uses MobileNetv2 network \cite{ref27} as the feature extractor for each branch of the dual-stream network to balance the inference speed and segmentation effect. For the dual-stream network, given an image $\mathit{I}_a\in\mathbb{R}^{3\times\mathit{H\times W}}$ and the corresponding optical flow estimation $\mathit{I}_m\in\mathbb{R}^{3\times\mathit{H\times W}}$, the dual-stream network extracts appearance features $\mathit{V}_{a,i}\in\mathbb{R}^{\mathit{C\times H\times W}}$ and motion features $\mathit{V}_{m,i}\in\mathbb{R}^{\mathit{C\times H\times W}}$ in the $i$-th stage, which is used to enhance the appearance features through the motion guidance module:
\begin{equation}
\label{equation:1}
U_{a,i} = F_{MG}(V_{a,i}, V_{m,i})
\end{equation}
where $\mathit{F_{MG}(\ast)}$ represents the motion guidance module, $\mathit{U}_{a,i}\in\mathbb{R}^{\mathit{C\times H\times W}}$ represents the enhanced $i$-stage appearance features. In particular, in order to preserve the details of shallow features, we don't set up the motion guidance module for the stage1 when we design the dual stream network. For the enhanced appearance features $\mathit{U}_{a,i}$ and the motion features  $\mathit{m}_{a,i}$ in stage $i$. After concatenating in the channel dimension, $\mathit{U}_i=\mathnormal{Concat}(\mathit{U}_{a,i},\mathit{U}_{m,i})\in\mathbb{R}^{2\mathit{C\times H\times W}}$ is sent to the multi-scale progressive fusion module to obtain the final segmentation map.

\subsection{Motion Guidance Module}
Co attention mechanism is widely used to extract the association information in different modal features. COSNet \cite{ref10} extracts the correlation information among features of multiple frames from the same video, and MATNet \cite{ref15} converts appearance features into motion attention representations through the co attention mechanism. The extensive application of co attention mechanism has the problem of huge computational cost while achieving good results, so improving the co attention mechanism can bring considerable efficiency improvement.

The naive co attention mechanism is shown in Fig. \ref{fig:2_1}. $\mathit{V}_a\in\mathbb{R}^{\mathit{C\times H\times W}}$ and $\mathit{V}_b\in\mathbb{R}^{\mathit{C\times H\times W}}$ are fed into the co attention module. Via compressing the channel of $1\times 1$ convolutional compression to $C/d$, then adjusting the dimension, we get $\overline{\mathit{V}}_a\in\mathbb{R}^{\mathit{C\times HW}}$ and $\overline{\mathit{V}}_b \in \mathbb{R}^{\mathit{C\times HW}}$. By calculating the similarity between the feature points in $\overline{\mathit{V}}_a$ and $\overline{\mathit{V}}_b$, the similarity matrix $\mathit{S}\in\mathbb{R}^{\mathit{HW}\times\mathit{HW}}$ is obtained. The matrix $\mathit{S}$ and its transposed matrix $\mathit{S}^{T}$ are multiplied with $\mathit{V}_a$ and $\mathit{V}_b$ respectively. After restoring the spatial dimension, obtaining the enhanced features $\mathit{U}_a\in\mathbb{R}^{\mathit{C\times H\times W}}$ and $\mathit{U}_b\in\mathbb{R}^{\mathit{C\times H\times W}}$.

This paper analyzes the advantages of co attention mechanism from the perspective of weighted summing. In co attention, $\mathit{I}_b\in\mathbb{R}^{\mathit{C\times 1\times 1}}$. It is obtained by summing each feature point in feature $\overline{\mathit{V}}_a$ weighted by a set of weights $\mathit{W}\in\mathbb{R}^{1\times\mathit{HW}}$, and this set of weights $\mathit{W}$ is normalized by the similarity matrix of the feature point $\mathit{I}_b$ at the corresponding position in $\overline{\mathit{V}}_b$ and all the feature points in $\overline{\mathit{V}}_a$. This method is similar to the Mul-tilayer perceptron (MLP), and the global calculation obtains the global receptive field. The difference is that the weights in MLP are learnable parameters, and the weights are defined similarly in co attention, which does not need to be learned, reducing the risk of overfitting.

The co attention mechanism obtains a global receptive field and avoids increasing the learnable parameters like MLP, and there is a problem with a sizeable computational amount. Similar to improving MLP by convolution, this paper uses the sliding window method to obtain local attention. Using the local instead of the global will significantly reduce the amount of computation. The performance of the model will not be degraded if the prior information of features is used properly.

Specifically, the similarity between each feature point $\mathit{I}_m$ in the motion feature $\mathit{V}_m$ and the feature points in the surrounding $K$ window is calculated, and the similarity matrix $\mathit{W}\in\mathbb{R}^{1\times\mathit{K}\times\mathit{K}}$ is obtained by normalizing. The feature point $\mathit{I}_a$ at the corresponding position in the appearance feature $\mathit{V}_a$ is weighted by the similarity matrix $\mathit{W}$ to sum the feature points in the $\mathit{K}$ window around it and the feature point $\mathit{I}_a$ of $\mathit{U}_a$ is obtained. In this way, the motion features obtain weighted weights by extracting semantic information through local attention and guide the learning of high-level semantics by passing weights to guide the weighting summation of appearance features.

The parallel computation of the motion guidance module implemented by the existing framework is shown in Fig. \ref{fig:2_2}. Appearance features $\mathit{V}_a\in\mathbb{R}^{\mathit{C}\times\mathit{H}\times\mathit{W}}$ expand according to $im2col$ and adjust the dimensions to get $\overline{\mathit{V}}_{a\_unfold}\in\mathbb{R}^{\mathit{K\times K\times C\times H\times W}}$. Motion features $\mathit{V}_m\in\mathbb{R}^{\mathit{C}\times\mathit{H}\times\mathit{W}}$ get through a layer of $1\times 1$ convolution compression channel to obtain $\overline{\mathit{V}}_m\in\mathbb{R}^{\mathit{(C/d)\times H\times W}}$, and expand $\overline{\mathit{V}}_m$ as $im2col$ and rearrange the dimensions yields $\overline{\mathit{V}}_{m\_unfold}\in\mathbb{R}^{\mathit{K\times K\times (C/d)\times H\times W}}$. By copying the $\overline{\mathit{V}}_m$ feature points $K\times K$ times and  repeating the new permutation dimension, $\overline{\mathit{V}}_{m\_repeat}\in\mathbb{R}^{\mathit{K\times K\times (C/d)\times H\times W}}$ is obtained. $\overline{\mathit{V}}_{m\_unfold}$ and $\mathit{V}_{m\_repeat}$ do similarity in the channel dimension to obtain similarity matrix $\mathit{S}\in\mathbb{R}^{\mathit{K\times K\times H\times W}}$. $\mathit{V}_{a\_unfold}$ multiplicate the similarity matrix $\mathit{S}$ to obtain the feature $\mathit{U}_a\in\mathbb{R}^{\mathit{C\times H\times W}}$.

\textit{Motion guidance module} in this paper is similar to convolution, except that the sliding window convolution weights defined by similarity are dynamic for each feature point in feature map, and do not need to be learned.

Compared with \textit{co attention module}, \textit{motion guidance module} greatly reduces the computation cost, and the computation quantity pair under different input sizes is shown in Table \ref{tabular:tb1}. At the same time, the motion guidance module can balance the model's ability to extract motion information with the ability to suppress background noise by limiting the maximum correlation distance (size $\mathit{K}$ of the sliding window). Bodywise, a $\mathit{K}$ that is too small cannot obtain enough exercise information. An excessively large $\mathit{K}$ increases the computation cost, and may extract the motion information of background objects similar to the foreground object. In particular, when $\mathit{K}=1$, the \textit{motion guidance module} degenerates into the motion feature $\mathit{V}_m$ and the appearance feature $\mathit{V}_a$ for element-by-element multiplication, which does not have the ability to obtain local attention in the motion feature $\mathit{V}_m$. Therefore, choosing the right $\mathit{K}$ is very important for the \textit{motion guidance module}. In addition to directly adjusting the value of $\mathit{K}$, we can also adjust the number of stacking layers of the module to simulate the effect of a larger $\mathit{K}$ value module, which further reduces the computation cost and improves the final effect.

\begin{table}[!t]
\renewcommand{\arraystretch}{1.3}
\caption{\label{tabular:tb1}\centering{Comparison of FLOPs of different modules}}
\centering
\resizebox{\linewidth}{!}
{\begin{tabular}{ccc}
\hline
Input Size & Co Attention Module  & Motion Guidance Module \\
\hline
$64\times 64\times 16$   &   10.0M   &   2.3M  \\
$64\times 64\times 32$   &   153.1M   &   9.0M \\
\hline
\end{tabular}}
\end{table}

\subsection{Multi-scale Progressive Fusion Module}
In different stages of \textit{dual-stream networks}, the extracted features have different resolution rates and contain different levels of semantic information, so it is particularly important to use these features rationally. Previous studies adopted the U-Net \cite{ref28} method of upsampling fusion strategy, and \textit{Atrous Spatial Pyramid Pooling} (ASPP) was used to enlarge receptive field of each stage, but the differences in semantic fusion of features at different stages were ignored.

\begin{figure}
	\centering
    \includegraphics[width=0.45\textwidth]{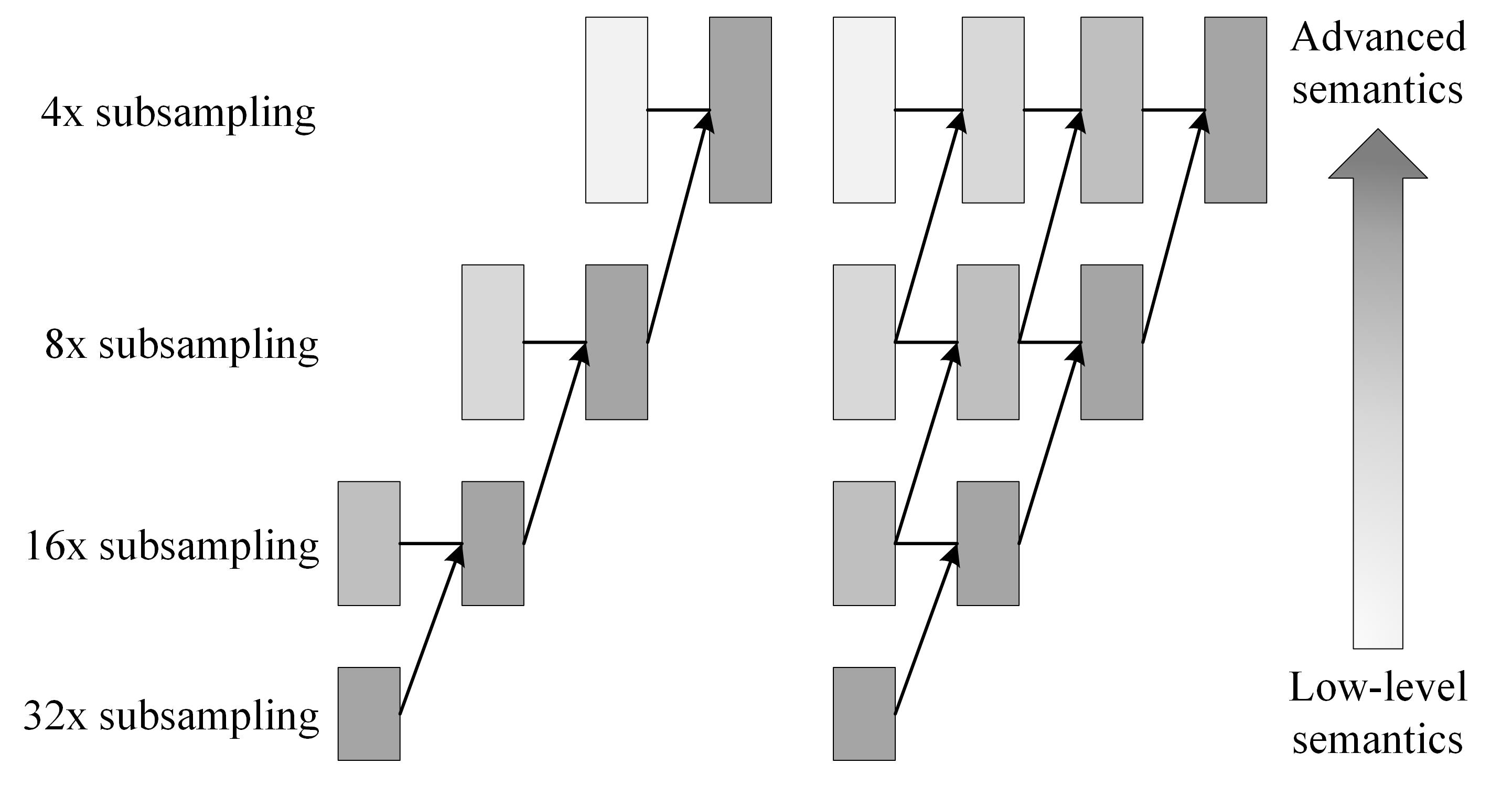}
	\caption{\centering{UNet style-like upsample module and multi-scale progressive fusion module}}
	\label{fig:3}
\end{figure}

As shown in Fig. \ref{fig:3}, the semantic information contained in the segmentation result graph can be regarded as a subset of high-level semantics, and low-resolution deep semantic features are combined with high-resolution shallow semantic features to obtain high-resolution high-level semantic features. However, with the continuous fusion progress, the semantic gap between the low-res features to be fused and the high-res features to be fused will widen, which is not conducive to the learning of fusion weights and reduces the segmentation performance.

Therefore, this paper proposes a \textit{Multi-scale Progressive Fusion Module} that adopts the strategy of continuously fusing high-level semantics into high-resolution features. The multi-stage features extracted by the \textit{dual-stream network} are respectively sent to the multi-stage branches processing different resolution features, then the low-res features are fused with the high-res features between each stage.

Specifically, for the $i$-branch feature $\mathit{U}_{j-1,i}\in{\mathbb{R}^{2\mathit{C}\times{(\mathit{H}/2)}\times{(\mathit{W}/2)}}}$ in the $j$-1th stage, first upsampling by 2 times is performed, and then stitched with the $i$-1 tributary feature $\mathit{U}_{j-1,i-1}\in{\mathbb{R}^{\mathit{C}\times\mathit{H}\times\mathit{W}}}$ in the channel dimension, and then sent to a two-layer residual structure to adjust the number of channels and fused, and finally obtain the $i-1$ features $\mathit{U}_{j,i-1}\in{\mathbb{R}^{\mathit{C}\times\mathit{H}\times\mathit{W}}}$ in the $j$th stage.
\begin{equation}
\label{equation:2}
U_{j,i-1}=F_{Conv}(Concat(U_{j-1,i-1},Up(U_{j-1,i})))
\end{equation}
\begin{equation}
\label{equation:3}
F_{conv}(*)=F_{res}(F_{res}(*))
\end{equation}

In this way, the semantic gap between fusion features is reduced, and improving the final segmentation result.

\section{Experiments}
\subsection{Training Phase}
In this paper, the MobileNet v2 \cite{ref27} network pre-trained using the ImageNet dataset is used as a feature extractor of the dual-stream network, and the Binary CrossEntropy Loss (BCE) is applied as the loss function when training. The training data is divided into two parts: Youtube VOS \cite{ref13} and DAVIS-16 \cite{ref12} dataset. Because the Youtube VOS \cite{ref13} dataset actually has category labels, which is not conducive to the training of category-independent video segmentation tasks, what's more, it has lower accuracy compared with DAVIS-16 dataset in segmentation annotations. As a consequence, this paper chooses to utilize Youtube VOS dataset for pre-training model and fine-tune the model using DAVIS-16 dataset for testing. At the same time, in order to ensure fairness, 9K training images are obtained by interval frame extraction on the Youtube VOS dataset, plus 2K training images of the DAVIS-16 dataset, for a total of 11K training images, which is the same as the training set of other algorithms. This paper uses the PWCNet network commonly used in previous research to preprocess the data set to obtain the optical flow estimation image.

This paper uses a common data augmentation strategy. For each training image, after random flipping, the image is rotated at a random angle between -$10^{\circ}$ and $10^{\circ}$ and then cropped and scaled to a size of $384\times672$. Both the fine-tuning and pre-training stage of network are ultilizing \textit{Stochastic Gradient Descent} (SGD) optimizer to optimize network parameters. We suppose that $\alpha$ represents learning rate, and $\alpha_1 = 1e-4$ is used for the feature extractor and \textit{motion guidance module}, and $\alpha_1 = 1e-3$ is used for the \textit{multi-scale progressive fusion module}, and $\alpha$ attenuation rate $\alpha_\downarrow = 0.9$, and weight decay rate $w_\downarrow = 5e-4$, and the batch size is 10 images per batch. Pre-training iterations 25 rounds, fine-tuning iterations 10 rounds. Build the network using the PyTorch 1.6.0 framework, and train and test the model on a Ge-Force GTX 2080 Ti GPU.

\subsection{DataSets}
In this paper, the model performance is tested on three datasets, DA-VIS-16 dataset \cite{ref12}, FBMS dataset \cite{ref3} and ViSal dataset \cite{ref25}. The DAVIS-16 dataset includes 50 videos, where 30 videos as training set and others as testing set. The FBMS dataset includes 59 videos, where 29 videos as training set and others as testing set, using a sparse labeling strategy of labeling one frame every 20 frames. The ViSal dataset consists of 17 test videos with a total of 193 frames of labeled images.

\subsection{Evaluation Metrics}
We adopt regional similarity $\mathit{R}$ and contour accuracy $\mathit{F}$ as standard evaluation metrics for the unsupervised video segmentation task in this paper. where $\mathit{F}$ is the union ratio of segmentation result and the labeled truth mask:
\begin{equation}
\label{equation:4}
R=\frac{|M\cap GT|}{|M\cup GT|}
\end{equation}
where $M$ represents segmentation result of the model and $GT$ represents ground-truth. $F$ treats the mask as a collection of closed contours, and computes the contour-based F-measure as follows:
\begin{equation}
\label{equation:5}
F=\frac{2Pre\times Rec}{Pre + Rec}
\end{equation}
where $Pre$ represents precision rate and $Rec$ represents recall rate. In addition, this paper also adopts the comprehensive  metric $\mathit{R}\&\mathit{F}$, which is expressed as the mean of both:
\begin{equation}
\label{equation:6}
R\&F=\frac{R+F}{2}
\end{equation}
The \textit{Mean Absolute Error} (MAE) and F-measure as evaluation metrics to evaluate the model for the video saliency detection task in this paper. The MAE describes direct pixel-level comparison of the binary saliency map with true map, as follows:
\begin{equation}
\label{equation:7}
MAE=\frac{1}{h\times w}\sum^{h}_{i=1}\sum^{w}_{j=1}\parallel{S_{i,j}-G_{i,j}}\parallel
\end{equation}
Where $S$ represents the binary significance graph, $G$ represents the true graph, $h$ (height) and $w$ (width) correspond to the size of image. F-measure is an evaluation metrics of complete accuracy and recall, which comprehensively reflects the performance of the algorithm, as follows:
\begin{equation}
\label{equation:8}
F_\beta=\frac{(1+{\beta}^2)Pre\times Rec}{{\beta}^2Pre+Rec}
\end{equation}
Where the weighted harmonic parameter $\beta^2$ is often set to 0.3.

\subsection{Results}
\begin{table}[!t]
\renewcommand{\arraystretch}{1.3}
\caption{\label{tabular:tb2}\centering{Evaluation results of different models on DAVIS-16 and FBMS dataset}}
\centering
\setlength{\tabcolsep}{5mm}
\resizebox{\linewidth}{!}
{\begin{tabular}{ccccc}
\hline
    &   \multicolumn{2}{c}{DAVIS-16} & \multicolumn{2}{c}{FBMS}   \\ 
\hline
Method & $R\&F$ & $R$ & $F$ & $R$ \\
\hline
LMP \cite{ref18}  & 68.0 & 70.0 & 65.9 & $-$      \\
LVO \cite{ref17} & 74.0 & 75.9 & 72.1 & $-$       \\
PDB \cite{ref14} & 75.9 & 77.0 & 74.5 & 74.0    \\
MBNM  \cite{ref19}& 79.5 & 80.4 & 78.5 & 73.9    \\
AGS \cite{ref20}& 78.6 & 79.7 & 77.4 & $-$       \\
COSNet \cite{ref10} & 80.0 & 80.5 & 79.4 & 75.6  \\
AGNN \cite{ref8} & 79.9 & 80.7 & 79.1 & $-$     \\
AnDiff \cite{ref21}& 81.1 & 81.7 & 80.5 & $-$    \\
MATNet \cite{ref15}& 81.6 & 82.4 & 80.7 & \textbf{76.1}  \\
Ours   & \textbf{83.6} & \textbf{83.7} & \textbf{83.4} & 75.9 \\
\hline
\end{tabular}}
\end{table}

On DAVIS-16 and the FBMS dataset, Table \ref{tabular:tb2} compares the performance of the proposed algorithm and several other advanced algorithms. In the DAVIS-16 dataset, $R$, $F$, and $R\&F$ are for references, and $R$ is for a reference on the FBMS dataset. This paper uses the $R$, $F$, and $R\&F$ as a reference on the DAVIS-16 dataset, simultaneously $R$ is used on the FBMS dataset. This paper does not use any post-processing methods other than flipping, such as the CRF post-processing method used in COSNet \cite{ref10} and MATNet \cite{ref15}. The method ranked on the DAVIS-16 dataset with $R\&F=83.6\%$, and with $R=75.9\%$ on the FBMS dataset, only to MAT-Net with a difference of 0.2\%. our method achieves a better results in the DAVIS-16 dataset are mainly due to the local attention of the motion guidance module, which suppresses a large amount of background noise, and the multi-scale progressive fusion module cooperates with a relatively large input resolution to get better segmentation results. Meanwhile, it is noted that the metric of the proposed algorithm on DAVIS-16 dataset is significantly higher than FBMS dataset, which is due to the poor effect of the PWCNet on FBMS dataset, and the better appearance features of motion information guidance cannot be obtained.
\begin{table}[!t]
\renewcommand{\arraystretch}{1.1}
\caption{\label{tabular:tb3}\centering{Evaluation results of different models on DAVIS-16, FBMS and ViSal dataset}}
\centering
\resizebox{\linewidth}{!}
{\begin{tabular}{ccccccc}
\hline
    &   \multicolumn{2}{c}{DAVIS-16} & \multicolumn{2}{c}{FBMS} & \multicolumn{2}{c}{ViSal}   \\ 
\hline
Method  &     MAE   &   $\mathit{F_m}$  &   MAE    &   $\mathit{F_m}$   &   MAE     &  $\mathit{F_m}$  \\
\hline
FCNS \cite{ref23}    &   .053   &   72.9    &   .100   &   73.5    &   .041   &   87.7    \\
FGRNE \cite{ref24}  &   .043    &   78.6    &   .083    &   77.9    &   .040    &   85.0    \\
TENET \cite{ref22}   &   .019    &   90.4    &   \textbf{.026}    &   \textbf{89.7}    &   \textbf{.014}    &   \textbf{94.9}    \\
MBNM \cite{ref19}    &   .031    &   86.2    &   .047    &   81.6    &   .047    &   $-$     \\
PDB \cite{ref14}    &   .030    &   84.9    &   .069    &   81.5    &   .022    &   91.7    \\
AnDiff \cite{ref21}  &   .044    &   80.8    &   .064    &   81.2    &   .030    &   90.4    \\
Ours    &   \textbf{.014}    &   \textbf{92.4}    &   .059    &   84.2    &   .019    &   92.1    \\
\hline
\end{tabular}}
\end{table}

\begin{table}[!t]
\renewcommand{\arraystretch}{1.3}
\caption{\label{tabular:tb4}\centering{Model parameters, FLOPs and infer latency of different methods}}
\centering
\setlength{\tabcolsep}{5mm}
\resizebox{\linewidth}{!}
{\begin{tabular}{cccc}
\hline
Method  &     COSNet \cite{ref9}  &   MATNet \cite{ref15} &   Ours  \\
\hline
Res.    &   473$\times$473   & 473$\times$473  &      384$\times$672 \\
$\#$Param(M)   &   81.2    &   142.7   &   6.4 \\
$\#$FLOPs(G)   &   585.5   &   193.7   &   5.4 \\
Latency (ms)    &   65  &   78  &   15  \\
\hline
\end{tabular}}
\end{table}

The video saliency detection task aims to achieve continuous extraction of motion-related saliency objects in video sequences by combining spatial-temporal information. Because of the similarity between UVOS tasks and video saliency detection, this paper also tests the video saliency detection indicators on the three datasets DAV-IS-16, FBMS, and ViSal, using MAE and F-m indicators as the basis, and the Table \ref{tabular:tb3} is shown the experiments results. It is obvious that our method obtains the best indicators on the DAVIS-16 dataset and competitive indicators on the FBMS dataset and ViSal dataset, proving the proposed method's effectiveness.

Since the algorithm in this paper selects the lightweight network and local attention module, in addition to the excellent performance on the standard data set, it also has advantages in the amount of model parameters and and the speed of model reasoning. Table \ref{tabular:tb4} compares model parameters, model computation, and inference delay between the proposed algorithm and two SOTA methods. The algorithm test does not use post-processing, and to eliminate the interference of different data loading methods on the inference speed of the model, this paper only tests the inference speed when the random matrix with the corresponding resolution and batch size is 1. First, the model reasoning 10 rounds of warm-up, then 60 rounds of statistical inference time, the highest and lowest 20 rounds of time are removed respectively, and the average time of the remaining 20 rounds is calculated to obtain the inference delay. The comparative experiments in Table \ref{tabular:tb4} show that the proposed algorithm effectively reduces the model parameters and computational cost, which has more advantages in practical applications. At the same time, the inference delay of this algorithm is only 15ms under higher-resolution input images. Compared with the MATNet \cite{ref15} method, which also uses motion features, the inference speed is increased by 5.2$\times$. Considering that the algorithm in this paper consumes less memory, it has more extensive concurrency on the same device.
\begin{figure*}[htbp]
    \centering
    \begin{subfigure}[t]{0.14\textwidth}
           \centering
           \includegraphics[width=\textwidth]{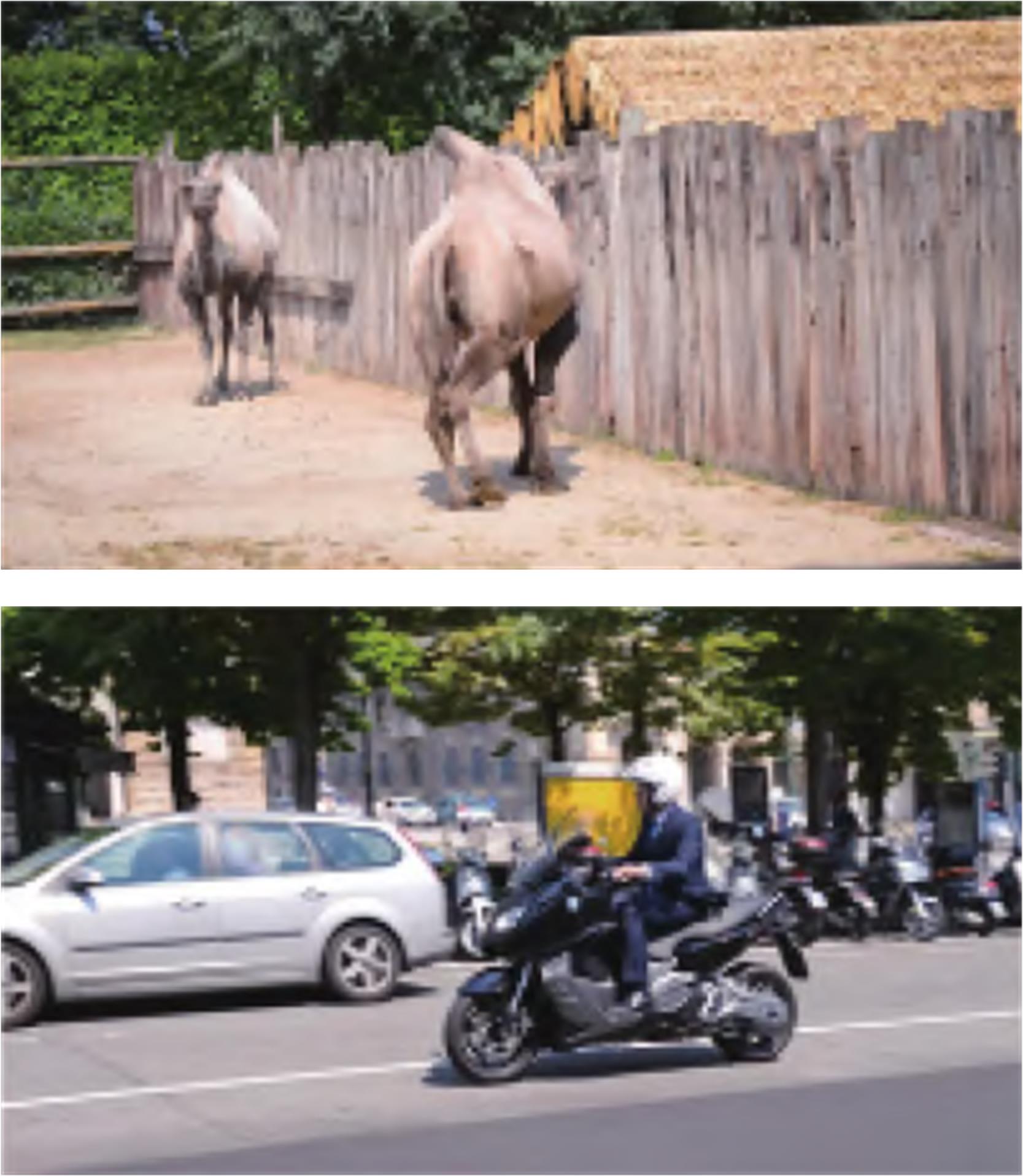}
           \caption{Image}
    \end{subfigure}
    \begin{subfigure}[t]{0.14\textwidth}
            \centering
            \includegraphics[width=\textwidth]{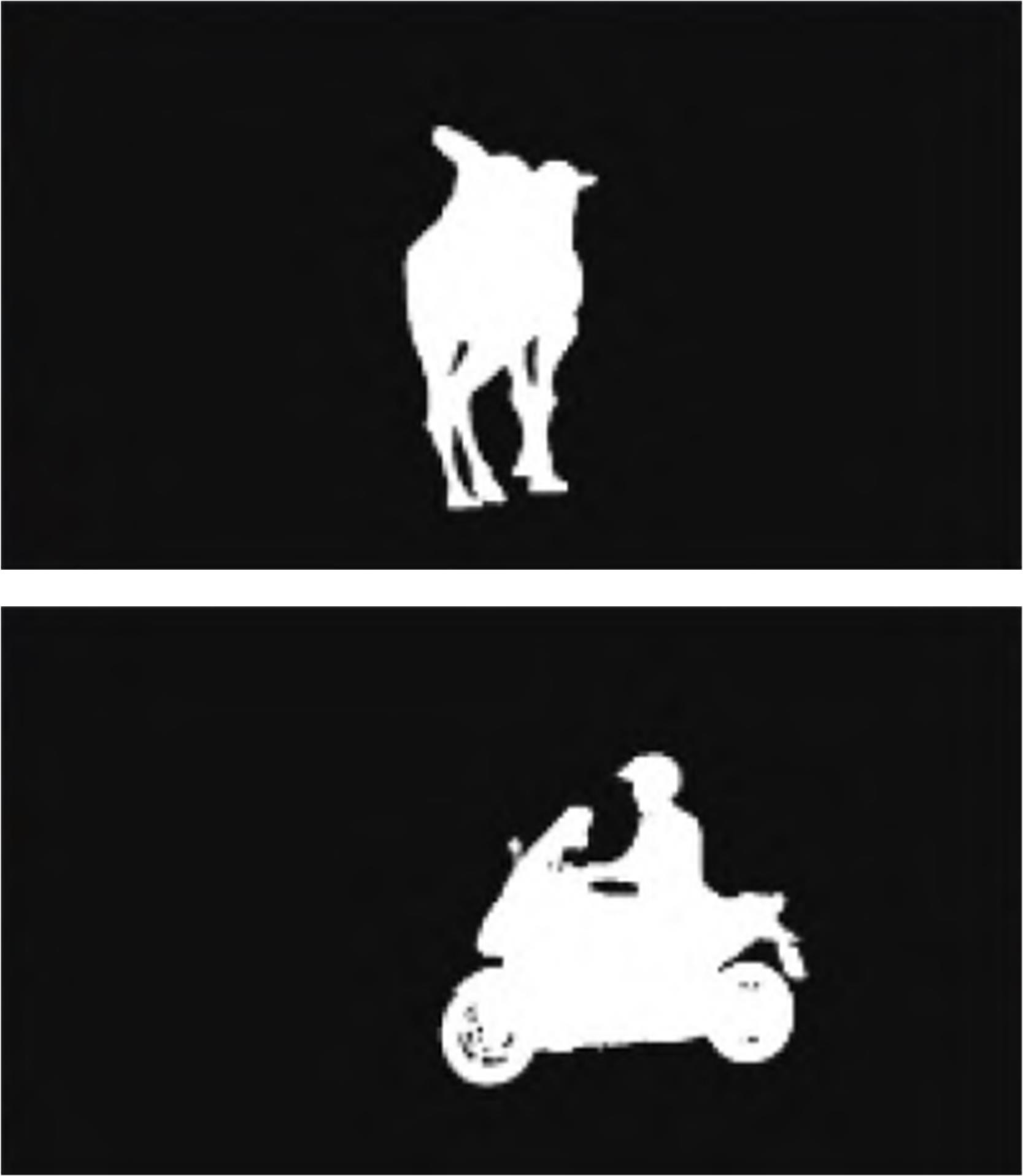}
            \caption{GT}
    \end{subfigure}
        \begin{subfigure}[t]{0.14\textwidth}
            \centering
            \includegraphics[width=\textwidth]{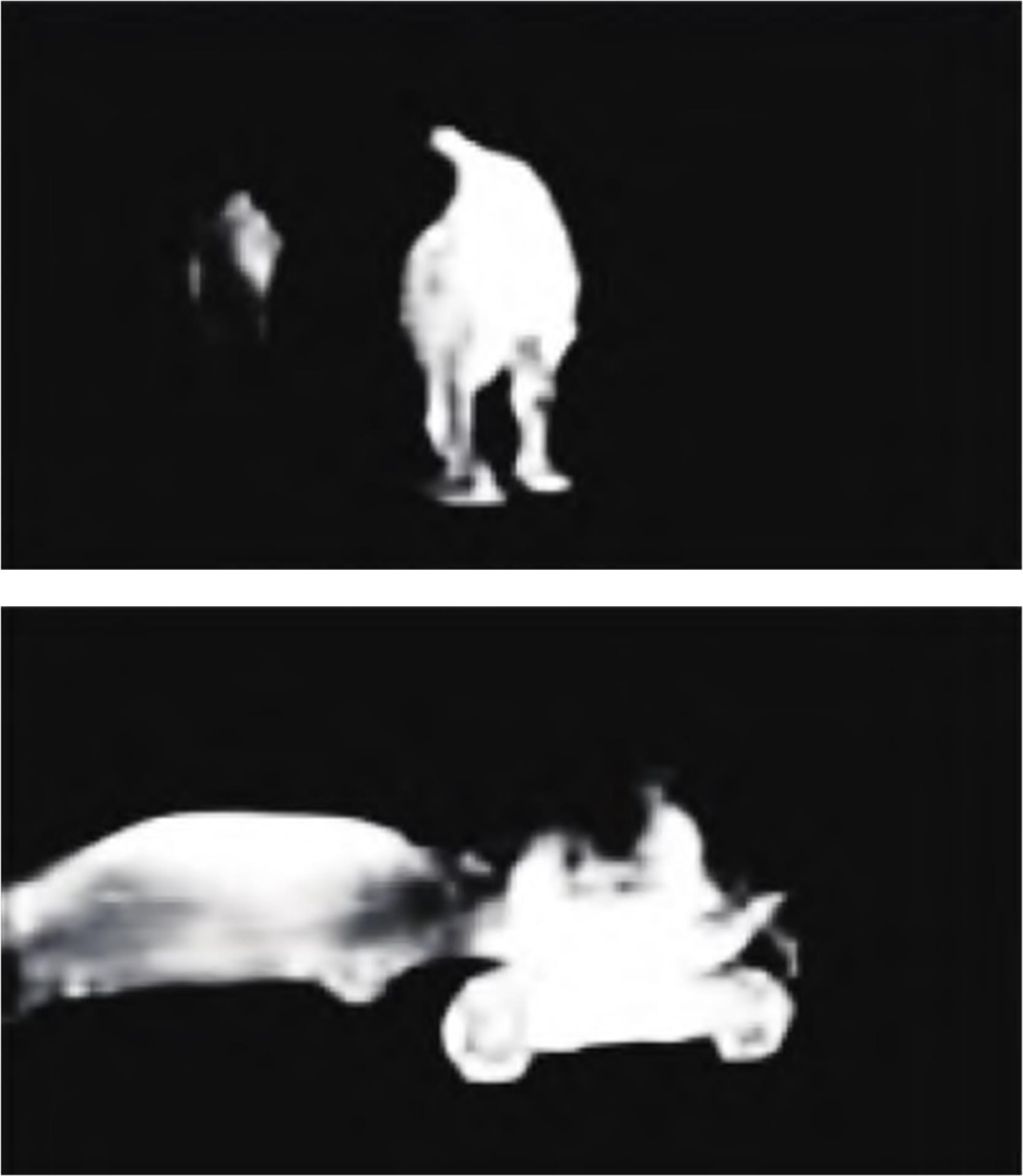}
            \caption{COSNet}
    \end{subfigure}
    \begin{subfigure}[t]{0.14\textwidth}
            \centering
            \includegraphics[width=\textwidth]{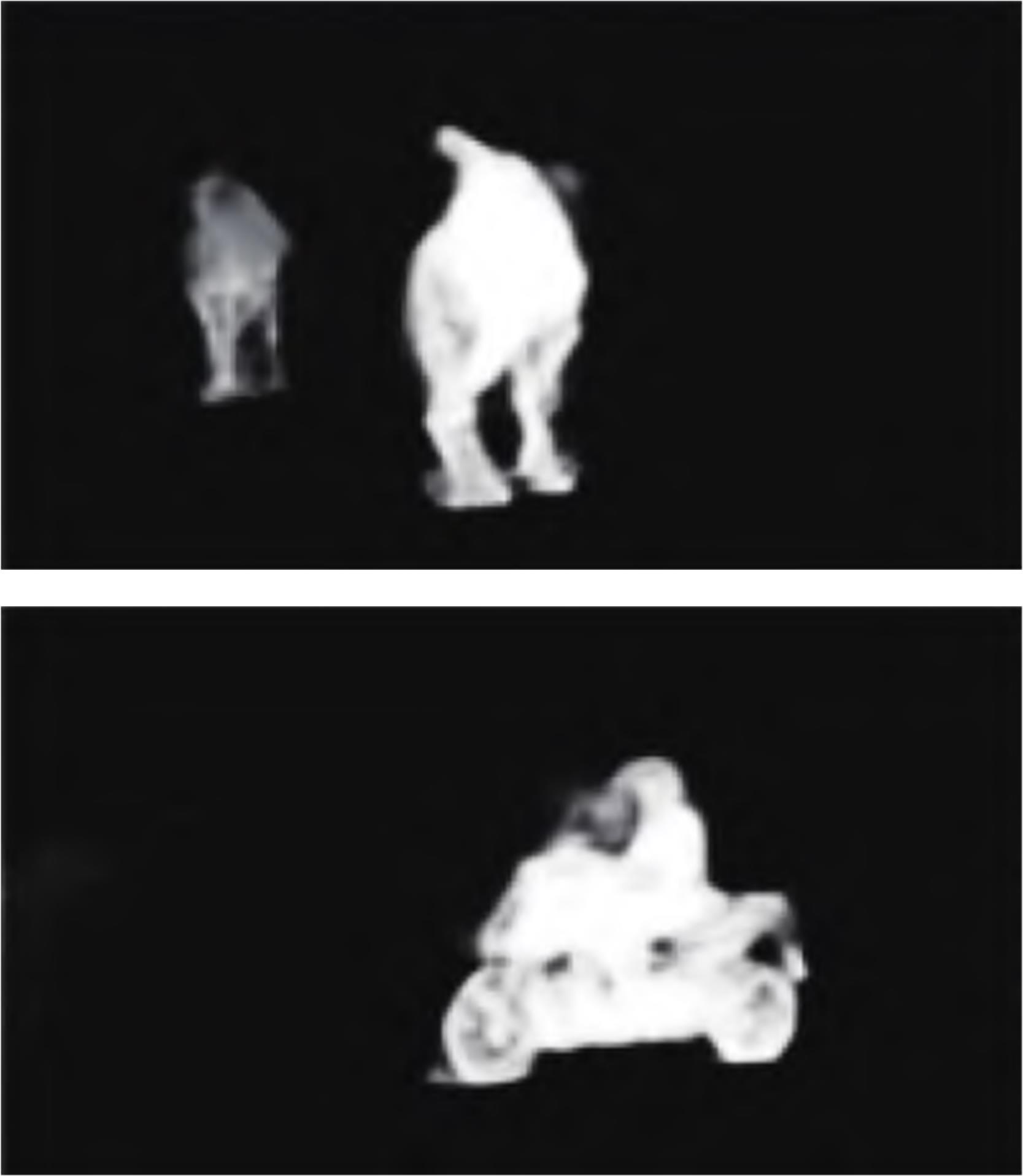}
            \caption{MATNet}
    \end{subfigure}
    \begin{subfigure}[t]{0.14\textwidth}
            \centering
            \includegraphics[width=\textwidth]{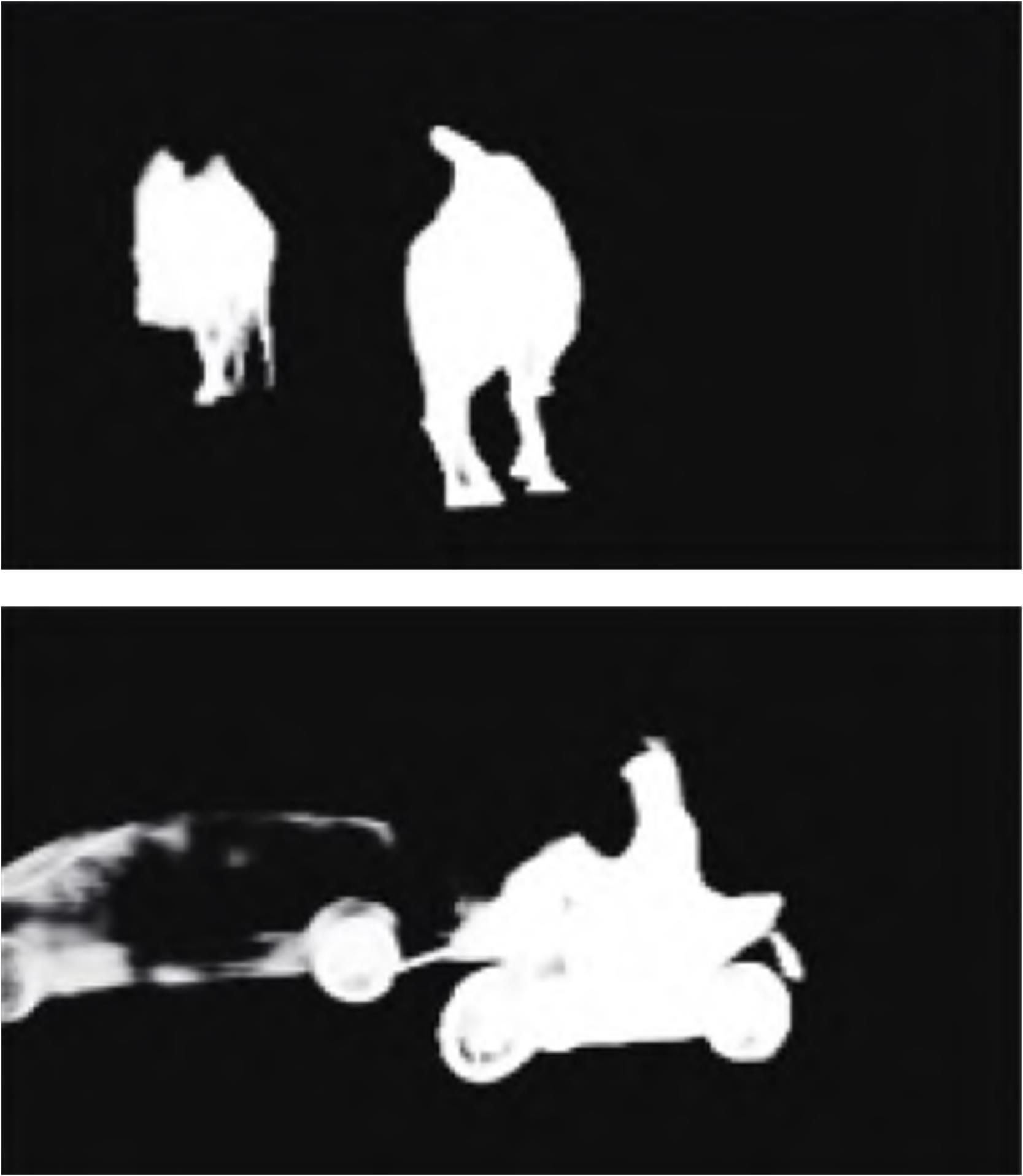}
            \caption{TENnt}
    \end{subfigure}
    \begin{subfigure}[t]{0.14\textwidth}
            \centering
            \includegraphics[width=\textwidth]{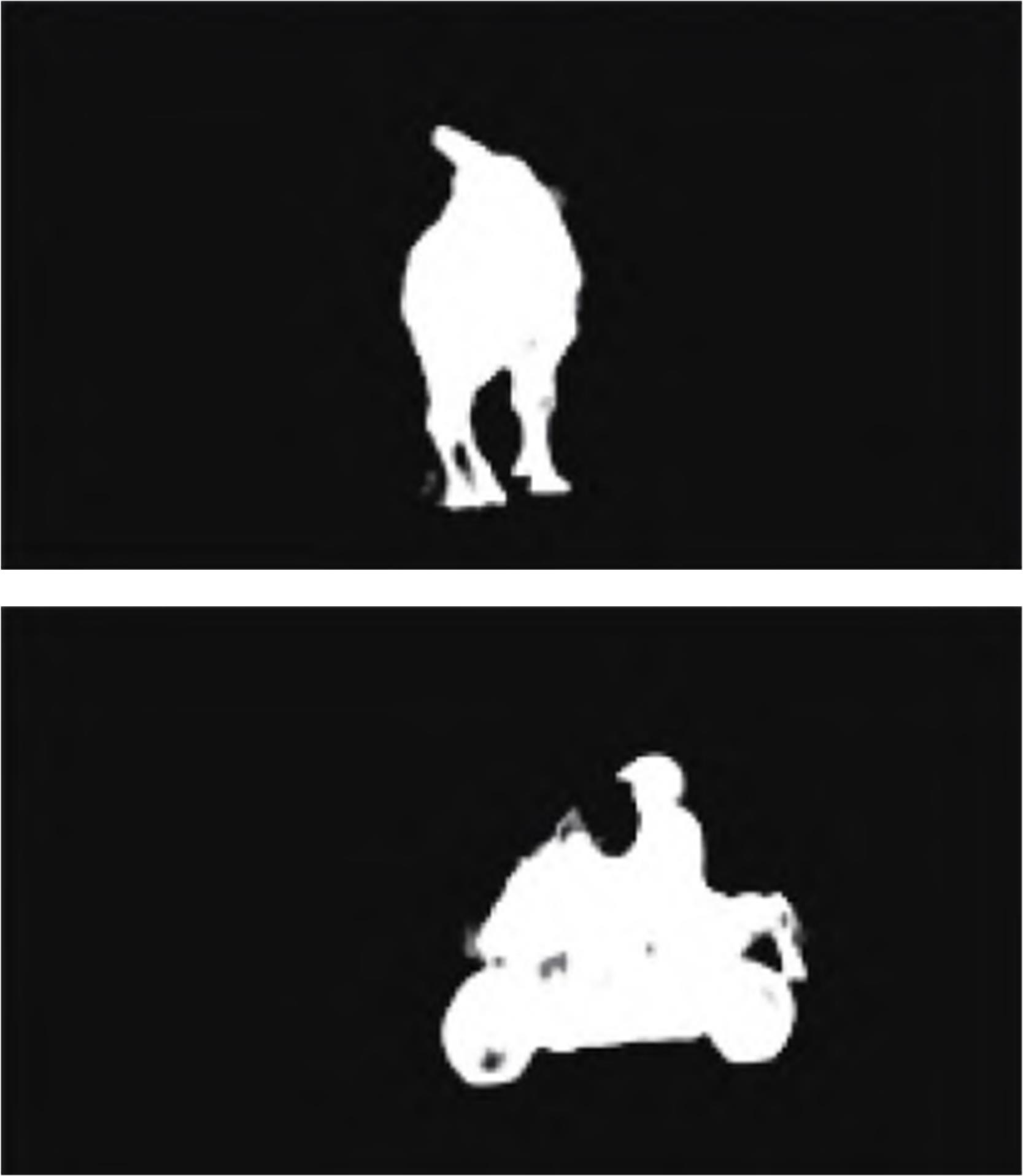}
            \caption{Ours}
    \end{subfigure}
    \caption{Comparative display of segmentation results\label{fig:4}}
\end{figure*}

\begin{table}[!t]
\renewcommand{\arraystretch}{1.1}
\caption{\label{tab5}\centering{Performance of different methods on GTX 2080 Ti}}
\centering
\setlength{\tabcolsep}{5mm}
\resizebox{\linewidth}{!}
{\begin{tabular}{cccc}
\hline
    &  Concurrency  &  FPS  &  Latency   \\ 
\hline
MATNet \cite{ref15}  &   18   &   16     &   62.4ms     \\
ours    &   \textbf{130}     &   \textbf{161}     &   \textbf{6.21ms}     \\
\hline
\end{tabular}}
\end{table}
For the sake of verify the efficiency of our method, the performance of  the method is tested and compared on the GeForce GTX2080 Ti GPU, and the results as shown in Table \ref{tab5}.

Benefiting from the lower amount of parameters and computation, the algorithm in this paper has a higher concurrency capability under the condition of making full use of 11G memory. It can simultaneously process 130 frames, which is 7.2$\times$ higher than MATNet. At the same time, the FPS of the algorithm in this paper reaches 161 frames per second, and the average inference delay is only 6.21ms.

Fig. \ref{fig:4} is shown the segmentation results of our method and other methods, and the comparison results prove that the algorithm in this paper can better suppress the background noise.

\subsection{Ablation Study}
\begin{table}[!t]
\renewcommand{\arraystretch}{1.2}
\caption{\label{tab6}\centering{Ablation experiments. $FG$ represents motion guidance and $U$ represents multi-scale progressive fusion}}
\centering
\setlength{\tabcolsep}{5mm}
\resizebox{\linewidth}{!}
{\begin{tabular}{cccc}
\hline
    &  Ours  &  $-FG-U$ &  $-FG$   \\ 
\hline
$R$  &   83.7   &   75.8      &   76.1     \\
$F$  &   83.4   &   73.5      &    75.6   \\
\hline
\end{tabular}}
\end{table}

This paper performs ablation experiment of the proposed algorithm on the DAVIS-16 dataset, using $R$ and $F$ as the primary metrics, and the experiment results as shown in Table \ref{tab6}. The baseline model extracts motion and appearance features based on the dual-stream network of MobileNet v2 network \cite{ref27} and uses point multiplication. Fusion semantics of motion and appearance feature matrix at each stage of the network, and finally obtains the segmentation results by UNet \cite{ref28} upsampling. The baseline model achieves only $R=75.8\%$ and $F=73.5\%$ results on the DAVIS-16 dataset. $F$ is improved by adding a multi-scale progressive fusion module ( $73.5\%$ $\to$ $75.6\%$ ). The segmentation performance has been dramatically improved by adding a motion guidance module. At the same time, this paper explores the best effect by inserting motion guidance modules with different parameters in a dual-stream network. It's can be seen from Table \ref{tab7}, the performance of model has achieved a significant improvement by adding the motion guidance module with $K=3$. Compared with the baseline model adding the multi-scale progressive fusion module, $R$ is improved by $6.7\%$, and F is improved by $6.8\%$.

At the same time, experiments show that the model performance rises and falls with the expansion of the $K$. This is mainly due to the increase of $K$, the movement information obtained by local attention, and the influence of background noise is also increasing. Finally, when $K=7$, a better balance is achieved.

\begin{table}[!t]
\renewcommand{\arraystretch}{1.2}
\caption{\label{tab7}\centering{Comparison of different kernel sizes and cascading times}}
\centering
\setlength{\tabcolsep}{5mm}
\resizebox{\linewidth}{!}
{\begin{tabular}{cccc}
\hline
 \multicolumn{2}{c}{Method}  &  \multicolumn{2}{c}{Ours}    \\ 
\hline
Kernel  &   Cascade     &   $R$     &       $F$     \\
3       &       1   &   82.8    &   82.4   \\
3   &   2   &   83.4    &       82.7    \\
3   &   3   &   83.7    &   83.4    \\
3   &   4   &   83.5    &   83.2    \\
5   &   1   &   83.2    &   82.6    \\
7   &   1   &   83.4    &   82.7    \\
9   &   1   &   83.1    &   82.4    \\
\hline
\end{tabular}}
\end{table}

Similar to using multiple layers of $3\times3$ convolutions to simulate more extensive convolutions, this paper also explores the impact of stacked motion guidance modules. By stacking two layers of modules with $K=3$, the effect of modules with $K=5$ is simulated, which reduces the computational cost and obtains better results. This paper attributes this to the fact that the module with $K=5$ only performs semantic extraction once. Replacing it with a similar module with $K=3$ can extract semantic information twice. The final performance exceeds that of the model with a larger $K$. At the same time, the stacked motion guidance module also appears that the performance rises and falls with the increase of $K$. Through Table \ref{tab7}, the algorithm of this paper selects the model of the motion guidance module with three layers of $K=3$ as the final model of this paper.

\begin{figure*}[htbp]
    \centering
        \includegraphics[width=0.9\textwidth]{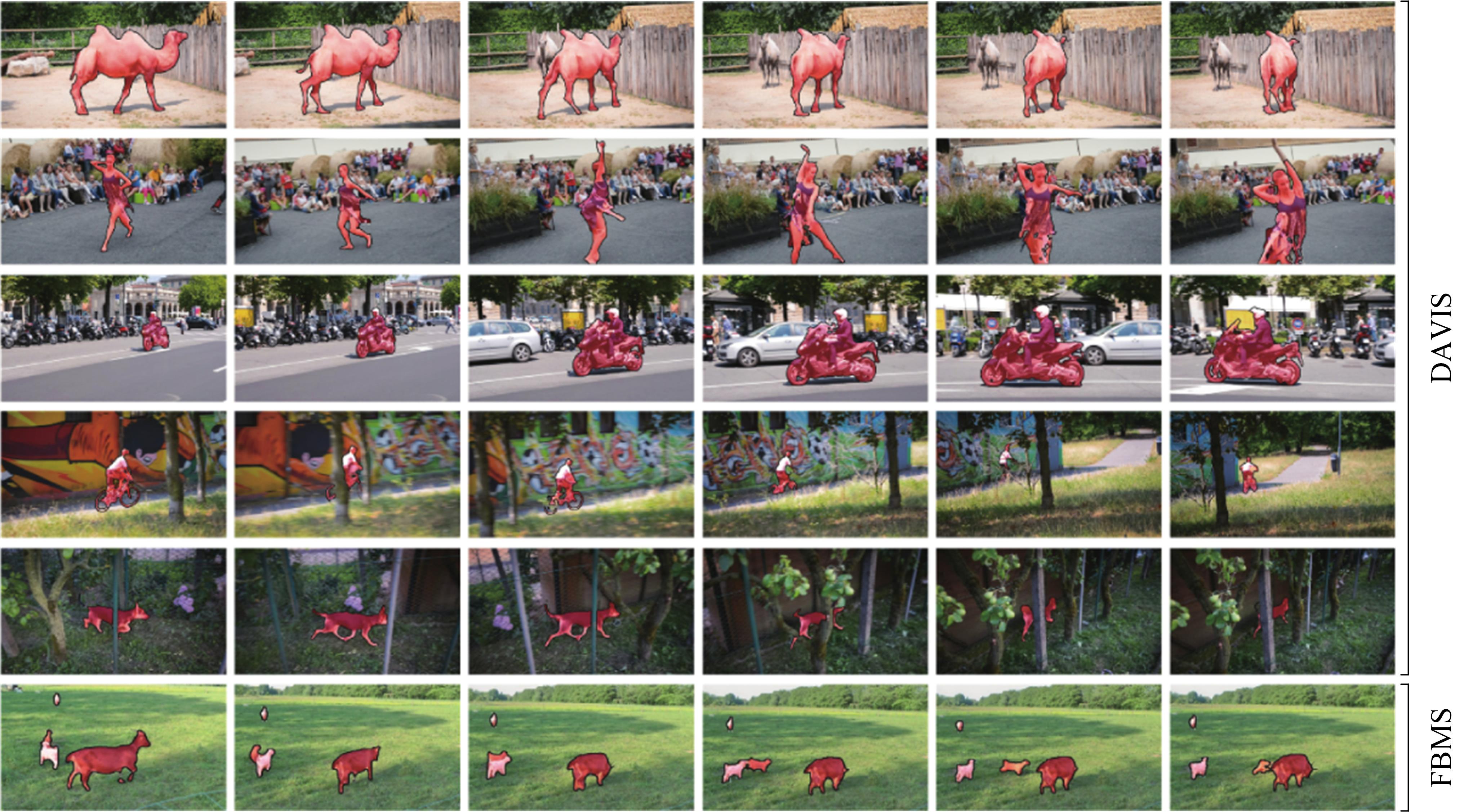}
        \caption{Display of segmentation results\label{fig:5}}
\end{figure*}

Fig. \ref{fig:5} shows that the algorithm has an excellent performance in various challenging scenarios. In the first line, this method can better distinguish the salient foreground from the similar objects in the background. The algorithm can accurately separate significant objects from the noisy background in the 2-nd and 3-rd lines. In the 4-th and 5-th rows, the algorithm can better handle the occlusion of objects. In the 6-th line, this article can be more advantageous to rationalize multiple significant prospect goals. The visual results demonstrate the availability of our algorithm.

\section{Conclusion}
This paper proposes an unsupervised video target segmentation method based on motion guidance, which first extracts motion and appearance characteristics through a dual-stream network. Then, the motion guidance module guides appearance features to learn apparent features to avoid enormous computational cost of the heavyweight feature extractor and the mutual attention mechanism. Finally, the multi-scale progressive fusion module continuously integrates high-level semantics into shallow features and obtains the final predicted segmentation results. The experimental results on multiple standard evaluation datasets thoroughly verify the superiority of the proposed method.


\begin{thebibliography}{1}

\bibitem{ref1}
Wenjun Zhu, Jun Meng, and Li Xu.
\newblock Self-supervised video object segmentation using integration-augmented attention.
\newblock Neurocomputing, vol. 455, pp. 325-339, 2021.

\bibitem{ref2}
Jing Liu, Jiaxiang Wang, Weikang Wang, and Yuting Su.
\newblock DS-Net: Dynamic spatiotemporal network for video salient object detection.
\newblock Digital Signal Processing, Vol. 130, 2022.

\bibitem{ref3}
Yuyan Dai, Canyan Zhu, Aiming Ji, and Lingfeng Mao.
\newblock A Synchronizing Method of Binocular Vision in Obstacle Detection.
\newblock In {\em Proceedings of the 2014 International Conference on Computer Network and Information Science},
  IEEE Computer Society, USA, pp. 1–4, 2014.
  
\bibitem{ref4}
Huong Ninh, and Guee-Sang Lee.
\newblock Adaptive Features for MRF based Cosegmentation.
\newblock In {\em International Conference on Ubiquitous Information Management \& Communication}, no. 40, pp. 1-5. ACM, 2016.

\bibitem{ref5}
Kaiping Wang, Yan Wang, Bo Zhan, Yujie Yang, Chen Zu, Xi Wu, Jiliu Zhou, Dong Nie, and Luping Zhou.
\newblock An Efficient Semi-Supervised Framework with Multi-Task and Curriculum Learning for Medical Image Segmentation.
\newblock International Journal of Neural Systems, Vol. 32, No. 09, 2022

\bibitem{ref6}
Hengxiang He, Yulong Qiao, Ximeng Li, Chunyu Chen, and Xingfu Zhang.
\newblock Optimization on multi-object tracking and segmentation in pigs' weight measurement.
\newblock Computers and Electronics in Agriculture,
Vol. 186, 2021.
  
\bibitem{ref7}
Xiankai Lu, Wenguan Wang, Jianbing Shen, David Crandall, and Jiebo Luo.
\newblock Zero-Shot Video Object Segmentation With Co-Attention Siamese Networks.
\newblock IEEE Transactions on Pattern Analysis and Machine Intelligence, vol. 44, no. 4, pp. 2228-2242, 2022.

\bibitem{ref8}
Tianfei Zhou, Jianwu Li, Shunzhou Wang, Ran Tao, and Jianbing Shen.
\newblock Matnet: Motion-attentive transition network for zero-shot video
  object segmentation.
\newblock {\em IEEE Transactions on Image Processing}, 29:8326--8338, 2020.

\bibitem{ref9}
Hui Wang, Weibin Liu, Weiwei Xing.
\newblock Video object segmentation via random walks on two-frame graphs comprising superpixels.
\newblock Journal of Visual Communication and Image Representation, vol. 80, 2021.
  
\bibitem{ref10}
Ziqin Wang, Jun Xu, Li Liu, Fan Zhu, and Ling Shao.
\newblock Ranet: Ranking attention network for fast video object segmentation.
\newblock In {\em Proceedings of the IEEE/CVF International Conference on
  Computer Vision}, pages 3978--3987, 2019.

\bibitem{ref11}
Nhat Hoang-Xuan, E-Ro Nguyen, Thuy-Dung Pham-Le, and Khoi Hoang-Nguyen.
\newblock Efficient One-Shot Video Object Segmentation.
\newblock In {\em 7th NAFOSTED Conference on Information and Computer Science (NICS)}, pp. 320-325, 2020.
  
\bibitem{ref12}
Mingqi Gao, Feng Zheng, James J. Q. Yu, Caifeng Shan, Guiguang Ding, and Jungong Han.
\newblock Deep learning for video object segmentation: a review.
\newblock Artificial Intelligence Review, 2022.

\bibitem{ref13}
Yadang Chen, Chuanyan Hao, Zhi-Xin Yang, and Enhua Wu.
\newblock Feelvos: Fast end-to-end embedding learning for video object
  segmentation.
\newblock Science China Information Sciences volume 65, 2019.

\bibitem{ref14}
Wenguan Wang, Shuyang Zhao, Jianbing Shen, Steven~CH Hoi, and Ali Borji.
\newblock Salient object detection with pyramid attention and salient edges.
\newblock In {\em Proceedings of the IEEE/CVF Conference on Computer Vision and
  Pattern Recognition}, pages 1448--1457, 2019.

\bibitem{ref15}
Ge-Peng Ji, Keren Fu, Zhe Wu, Deng-Ping Fan, Jianbing Shen, and Ling Shao.
\newblock Full-duplex strategy for video object segmentation.
\newblock In {\em Proceedings of the IEEE/CVF international conference on
  computer vision}, pages 4922--4933, 2021.

\bibitem{ref16}
J. Sarala Devi, and A. Razia Sulthana.
\newblock Video object segmentation guided refinement on foreground-background objects.
\newblock Multimedia Tools and Applications, 2022.

\bibitem{ref17}
Haofeng Li, Guanqi Chen, Guanbin Li, and Yizhou Yu.
\newblock Motion guided attention for video salient object detection.
\newblock In {\em Proceedings of the IEEE/CVF international conference on
  computer vision}, pages 7274--7283, 2019.

\bibitem{ref18}
Jordi Pont-Tuset, Federico Perazzi, Sergi Caelles, Pablo Arbel{\'a}ez, Alex
  Sorkine-Hornung, and Luc Van~Gool.
\newblock The 2017 davis challenge on video object segmentation.
\newblock {\em arXiv preprint arXiv:1704.00675}, 2017.

\bibitem{ref19}
Sucheng Ren, Wenxi Liu, Yongtuo Liu, Haoxin Chen, Guoqiang Han, and Shengfeng
  He.
\newblock Reciprocal transformations for unsupervised video object
  segmentation.
\newblock In {\em Proceedings of the IEEE/CVF conference on computer vision and
  pattern recognition}, pages 15455--15464, 2021.

\bibitem{ref20}
Mingmin Zhen, Shiwei Li, Lei Zhou, Jiaxiang Shang, Haoan Feng, Tian Fang, and
  Long Quan.
\newblock Learning discriminative feature with crf for unsupervised video
  object segmentation.
\newblock In {\em European Conference on Computer Vision}, pages 445--462.
  Springer, 2020.
 
\bibitem{ref21}
Lu Zhang, Jianming Zhang, Zhe Lin, Radom{\'\i}r M{\v{e}}ch, Huchuan Lu, and You
  He.
\newblock Unsupervised video object segmentation with joint hotspot tracking.
\newblock In {\em European Conference on Computer Vision}, pages 490--506.
  Springer, 2020.

\bibitem{ref22}
Miao Zhang, Jie Liu, Yifei Wang, Yongri Piao, Shunyu Yao, Wei Ji, Jingjing Li,
  Huchuan Lu, and Zhongxuan Luo.
\newblock Dynamic context-sensitive filtering network for video salient object
  detection.
\newblock In {\em Proceedings of the IEEE/CVF International Conference on
  Computer Vision}, pages 1553--1563, 2021.

\bibitem{ref23}
Yanwei Pang, Yazhao Li, Jianbing Shen, and Ling Shao.
\newblock Towards bridging semantic gap to improve semantic segmentation.
\newblock In {\em Proceedings of the IEEE/CVF International Conference on
  Computer Vision}, pages 4230--4239, 2019.

\bibitem{ref24}
Qiudan Zhang, Shiqi Wang, Xu Wang, Zhenhao Sun, Sam Kwong, and Jianmin Jiang.
\newblock A multi-task collaborative network for light field salient object
  detection.
\newblock {\em IEEE Transactions on Circuits and Systems for Video Technology},
  31(5):1849--1861, 2020.

\bibitem{ref25}
Min Hu, Ruimin Hu, Zhongyuan Wang, Zixiang Xiong, and Rui Zhong.
\newblock Spatiotemporal two-stream LSTM network for unsupervised video summarization.
\newblock Multimedia Tools and Applications, volume 81, 2022.

\bibitem{ref26}
Qianqian Zhou, Zuxiang Situ, Shuai Teng, Hanlin Liu, Weifeng Chen, and Gongfa Chen.
\newblock Automatic sewer defect detection and severity quantification based on pixel-level semantic segmentation.
\newblock Tunnelling and Underground Space Technology, Vol.  123, 2022.

\bibitem{ref27}
Weikuan Jia, Zhifen Wang, Zhonghua Zhang, Xinbo Yang, Sujuan Hou, and Yuanjie Zheng.
\newblock A fast and efficient green apple object detection model based on Foveabox.
\newblock Journal of King Saud University - Computer and Information Sciences, Vol. 34, no. 8, pp. 5156-5169, 2022.

\bibitem{ref28}
P. Celard, E. L. Iglesias, J. M. Sorribes-Fdez, R. Romero, A. Seara Vieira, and L. Borrajo.
\newblock A survey on deep learning applied to medical images: from simple artificial neural networks to generative models.
\newblock Neural Computing and Applications, 2022.

\bibitem{ref29}
Qing Zhang, Yanjiao Shi, Xueqin Zhang, and Liqian Zhang.
\newblock Residual attentive feature learning network for salient object detection.
\newblock Neurocomputing, Vol. 501, pp. 741-752, 2022.

\bibitem{ref30}
Rabi Sharma, Muhammad Saqib, C. T. Lin, and Michael Blumenstein.
\newblock A Survey on Object Instance Segmentation.
\newblock In {\em Proceedings of the IEEE/CVF conference on computer vision and
  pattern recognition}, pages 13906--13915, 2020.

\bibitem{ref31}
Alana de Santana Correia, and Esther Luna Colombini.
\newblock Attention, please! A survey of neural attention models in deep learning.
\newblock Artificial Intelligence Review, vol. 55, pp. 6037–6124 2022.

\bibitem{ref32}
Weitao Wan, Jiansheng Chen, Ming-Hsuan Yang, and Huimin Ma.
\newblock Co-attention dictionary network for weakly-supervised semantic segmentation.
\newblock Neurocomputing, Vol. 486, pp. 272-285, 2022.

\end{thebibliography}
\end{document}